\documentclass[11pt]{article}
\usepackage{booktabs}
\usepackage[table]{xcolor} %
\usepackage{svg}
\svgsetup{inkscapelatex=true} %
\svgpath{{images/}}  
\usepackage{makecell} 
\usepackage{multirow} %
\usepackage{amsmath} 
\usepackage{adjustbox}

\usepackage[final]{acl}
\usepackage{graphicx}
\usepackage{float}

\usepackage{times}
\usepackage{latexsym}

\usepackage[T1]{fontenc}

\usepackage[utf8]{inputenc}

\usepackage{microtype}

\usepackage{inconsolata}

\usepackage{graphicx}

\usepackage{cleveref}
\usepackage{enumitem}

\setlist{nosep,noitemsep}  %

\title{Robustness of Neurosymbolic Reasoners on First-Order Logic Problems}

\author{Hannah Bansal\textsuperscript{1,}\thanks{Work done while at The University of Melbourne.} \quad
Kemal Kurniawan\textsuperscript{2} \and Lea Frermann\textsuperscript{2} \\
\textsuperscript{1} School of Computing Technologies, RMIT University, Melbourne, Australia \\
\textsuperscript{2} School of Computing and Information Systems \\ The University of Melbourne, Melbourne, Australia \\
\texttt{hannah.bansal@rmit.edu.au} \\
\texttt{\{kurniawan.k,lea.frermann\}@unimelb.edu.au} \\
}

\begin{document}
\maketitle
\begin{abstract}

Recent trends in NLP aim to improve reasoning capabilities in Large Language Models (LLMs), with key focus on generalization and robustness to variations in tasks. Counterfactual task variants introduce minimal but semantically meaningful changes to otherwise valid first-order logic (FOL) problem instances altering a single predicate or swapping roles of constants to probe whether a reasoning system
can maintain logical consistency under perturbation.
Previous studies showed that
LLMs becomes brittle on
counterfactual variations, suggesting
that they often rely on spurious surface patterns
to generate responses.
In this work, we explore if a neurosymbolic (NS) approach that integrates an LLM and a symbolic logical solver could mitigate this problem. Experiments across LLMs of varying sizes show that NS methods are more robust but perform worse overall that purely neural methods. We then propose NSCoT that combines an NS method and Chain-of-Thought~(CoT) prompting and demonstrate that while it improves performance, NSCoT still lags behind standard CoT. Our analysis opens research directions for future work. The code for this work is available at \url{https://github.com/hannahhb/counterfactual_NS_eval}

\end{abstract}

\section{Introduction}

\begin{figure}[ht]
  \includegraphics[width=\columnwidth,clip,trim=0.5cm 0 0.5cm 0]{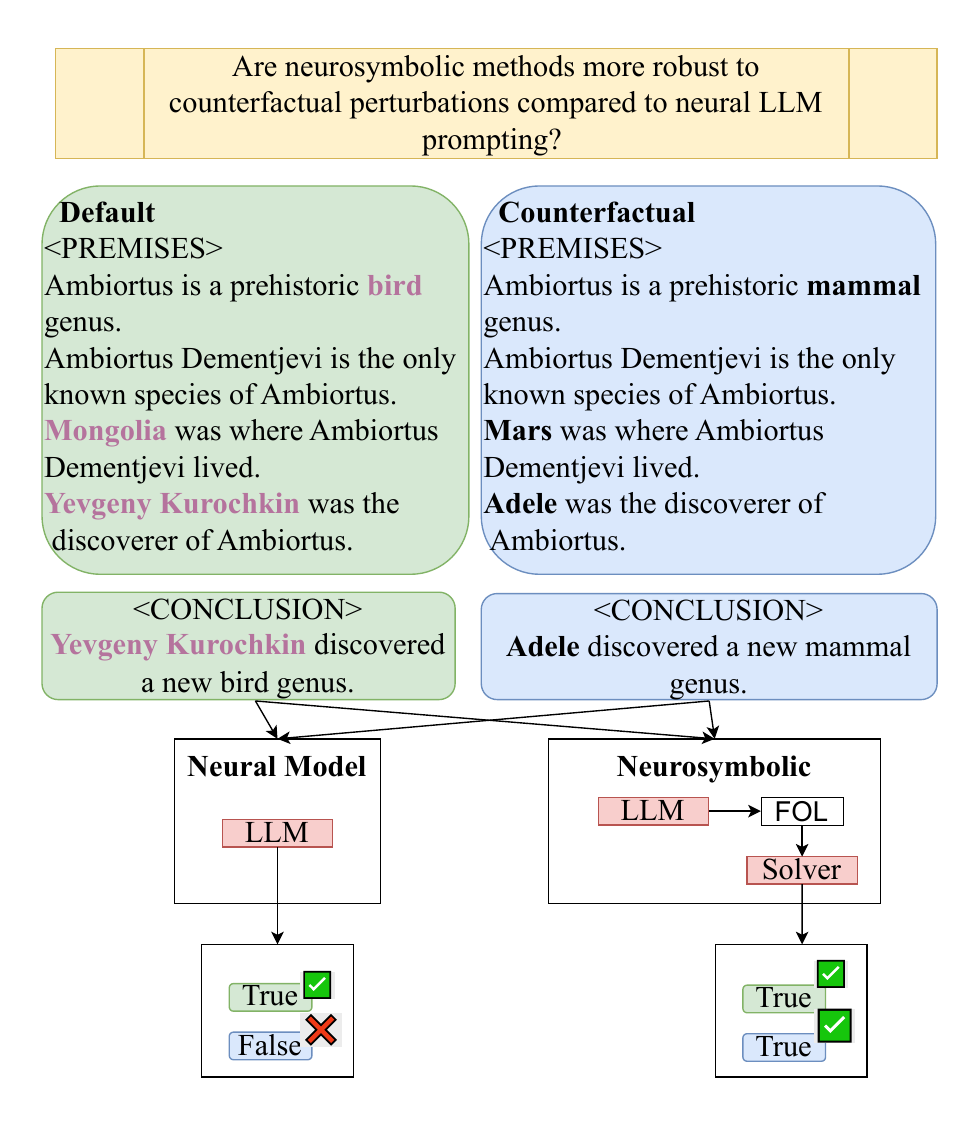}
  \caption{Illustration of our data and models. We test models in their ability to reason over default and counterfactual inputs, where key nouns were swapped (top). We compare fully neural models (LLMs) with neurosymbolic methods that combine LLMs with logical solvers. In our example the neural model fails on the counterfactual input but the neurosymbolic method makes correct predictions (bottom), suggesting higher robustness. Example taken from~\cite{wu2024reasoningreciting}.}
  \label{fig:cf-examples}
\end{figure}

LLMs have shown remarkable success on a range of different tasks including logical reasoning \citep{wei2022cot, deepseek2024v3}, mathematics~\citep{lewkowycz2022quantreasoning}, coding \citep{chen2021codeeval}, and creative tasks \citep{ramesh2021zeroshot}. These models have up to trillions of parameters and are pretrained to predict the most likely next word given the preceding words \citep{radford2018gpt} on vast amounts of digitized data
~\citep{brown2020fewshot, chowdhery2022palm}.
However, they cannot inherently perform formal rule based inference how a symbolic solver would.

\citet{wu2024reasoningreciting} showed that these models suffer in test conditions that systematically differ from the inputs they observed during training by leveraging ``counterfactual'' (CF) tasks.
CF tasks are carefully constructed to require the same reasoning as the original problem but with different assumptions. For example, for logical reasoning, \citet{wu2024reasoningreciting} demonstrated that replacing nouns and adjectives with less plausible alternatives in the input data to render the statements incompatible with any observed data can lead to a performance drop of 20\%.  This massive drop suggests that LLMs
memorize their training data rather than learning how to reason logically. \Cref{fig:cf-examples} illustrates this process.

In this work, we explore whether a neurosymbolic~(NS) approach could mitigate this problem for logical reasoning tasks. In such an approach, a neural network is used to translate natural language~(NL) premises and conclusions to first-order logic~(FOL) statements. Then, an FOL solver is used to determine if the conclusion logically follows from the premises.
Intuitively, delegating the reasoning process to an external tool---a symbolic FOL solver---should make the process less sensitive to ``counterfactual'' perturbations.

Our main research question is:
\emph{are NS methods more robust to counterfactual variation than purely neural approaches?} To answer this question, we employ LINC~\citep{olausson2023linc}, an NS method for logical reasoning that uses an LLM to translate NL into FOL, and compare against the LLM alone. Our experiments across LLMs with 7B to 32B parameters answer the question in the positive.
To the best of our knowledge, we are the first to test the sensitivity of NS methods on counterfactual tasks.

However, we also find that LINC performs worse overall.
We hypothesise that the model requires more explicit guidance
to correctly convert NL statements to FOL. To address this, we then propose NSCoT which combines Chain-of-Thought~(CoT) prompting \cite{wei2022cot} and NS approaches to improve the overall performance of NS methods. Specifically, we include example CoT reasoning in the LLM's in-context learning examples and prompt the LLM to generate its CoT reasoning when translating into FOL. We find that NSCoT substantially outperforms LINC but still lags behind a purely LLM approach with CoT.

In sum, our contributions are:
\begin{enumerate}
    \item We provide the first rigorous exploration of the robustness of neurosymbolic methods for logical reasoning on counterfactual inputs. We show that neurosymbolic methods are more robust but achieve lower performance than purely neural methods.
    \item We then propose NSCoT that integrates neurosymbolic methods and CoT. We demonstrate that NSCoT outperforms standard neurosymbolic methods, but still lags behind CoT.
\end{enumerate}

\section{Related work}
\subsection{Testing LLMs on perturbed data}

Prior work has studied the sensitivity of LLMs to data perturbations. \citet{jiang2024tokenbias} demonstrated that simply replacing the primary noun in the prompt~(e.g., from “Linda” to “Bob”)
causes the model to fail, despite the logical structure of the task remaining unchanged. The paper concluded that LLMs suffer from the \textit{token bias problem}, a phenomenon wherein LLMs exhibited a disproportionate reliance on frequently occurring lexical items such as specific nouns or structural cues to guide their reasoning process. \citet{wu2024reasoningreciting} introduced perturbations for a variety of tasks including arithmetic, code execution, logic, drawing, chord fingering, and chess. Their perturbations make the tasks deviate from
the default, generally assumed conditions, which they called as ``counterfactual''~(CF). These CFs were manually constructed and carefully controlled to fix the difficulty levels of items and keep comparisons fair. They hypothesised that LLMs simply memorise their training examples rather than actually reasoning about problems. They found that although CoT reasoning and few-shot learning can reduce the gap in performance between default and CFs, a significant performance drop remains.

The token bias problem has also been studied in the mathematical domain.
The GSM-Symbolic benchmark by~\citet{mirzadeh2024gsmsymbolic} systematically tests the impact of token bias by creating parsable templates and sampling different proper names and numerical values in mathematical problems. They showed that compared to default settings, numerical perturbations lead to about a 4\% performance drop.
A similar effect is found with proper names, further showing that the accuracy difference compounds when combined with the numerical perturbations.

Prior work has also shown that even when tasks remain within the reasoning capacity of humans, LLMs exhibit significant failures when problem complexity increases, presumably because such problems are rare in their pretraining data. For example, \citet{apple2025illusion} used the Tower of Hanoi problem as an example where they showed that while LLMs solve the problem with a small number of disks, their reasoning fails with larger number of disks.

Similar to prior work, we apply LLMs to a perturbed dataset to test their sensitivity to data perturbations. In contrast, however, we use LLMs in a neurosymbolic approach where we delegate the reasoning step to a symbolic solver. We note that in the literature, terms such as “counterfactual” are also used to describe hypothetical conditions that are false in the real world~\citep{li2023counterfactual}.
We use the term “counterfactual” hereinafter to be consistent with \citet{wu2024reasoningreciting}, i.e., perturbed data samples.

\subsection{Neurosymbolic reasoning}

Recent work has explored the integration between LLMs and symbolic systems to improving the reasoning capabilities of LLMs. Such neurosymbolic methods introduce a two-stage pipeline where natural language is first translated into FOL statements, which are then passed into a symbolic solver for resolution. This positions the LLM to perform a more abstract role of semantic parsing rather than direct reasoning. For logical reasoning, recent neurosymbolic methods include LINC~\citep{olausson2023linc}, Logic-LM~\citep{pan2023logiclm}, and SatLM~\citep{ye2023satlm}.

In this work, we use LINC as a representative of neurosymbolic reasoning approaches and test its robustness to counterfactual perturbations.
To the best of our knowledge, we are the first to test LINC
in the context of counterfactual examples.

\section{Methods}
\subsection{Dataset}
We mainly work with the data from~\citet{wu2024reasoningreciting}, a subset of the FOLIO dataset~\cite{han-etal-2024-folio} which has been turned into a `counterfactual' data set. In FOLIO, `premises' are paired with different `conclusions' which either logically follow from the premises (True), or they don't (False), or a conclusion cannot be drawn given the information in the premises (Uncertain). The task of a model is to classify the given the natural language premises and conclusion into one of these 3 labels.

\citet{wu2024reasoningreciting} manually swapped core noun variables in the premises with semantically implausible nouns which however do not alter the logical conclusion. Figure \ref{fig:cf-examples} shows an example. Intuitively, a robust reasoner would not be confused by this, while a brittle reasoning model which relies on surface cues would. There are 81 examples in this dataset, which we will refer to as RR~(Reasoning or Reciting, from \citeauthor{wu2024reasoningreciting}'s paper title). 

Due to limited examples and low representation of more complex reasoning problems in RR we also compare the performance of our methods on the full FOLIO validation set of 204 samples, even though this does not have a counterfactual variant. We note that this is an in-distribution task since we pass examples from the train split of the same dataset as in-context learning input to the LLM.

\subsection{Neurosymbolic Methods}

As our neurosymbolic method, we use LINC~\citep{olausson2023linc} where an LLM acts as a semantic parser to translate natural language premises and conclusions into FOL statements. These statements are then passed into a logic solver called Prover9~\citep{mccune2024prover9} to predict the classification label.
We use 8 in-context learning examples following \citet{olausson2023linc}.

The solver raises an error if an input cannot be parsed~(i.e., if the LLM generate FOL statements that do not comply with Prover9's format).
To handle this, we follow \citet{olausson2023linc}: we prompt the LLM 10 times to obtain 10 generations, pass each of them to Prover9 to get a predicted label, and perform majority voting to get the final predicted label excluding the error cases. If all generations are errors, we count the prediction as wrong in performance evaluation.

\begin{table*}[t]\small
\centering
\begin{tabular}{llccccp{0.1cm}|c}
\toprule
\textbf{Model} &  & \textbf{Naïve} & \textbf{ScratchPad} & \textbf{CoT} & \textbf{LINC} && \textbf{NSCoT} \\
\midrule
\multirow{3}{*}{Mistral0.3 7B} 
  & Default & 85.19 & \textbf{87.65} & \textbf{87.65} & 60.49 && 54.32 \\
  & CF      & 44.44 & \textbf{65.43} & 61.73 & 56.79 && 49.38 \\
  & $\Delta$   & -40.75$^*$ & -21.99$^*$ & -25.92$^*$ & \textbf{-3.70} && -4.94 \\
\midrule
\multirow{3}{*}{Qwen2.5 7B} 
  & Default & 83.95 & 86.42 & \textbf{87.65} & 66.67 && 62.96   \\
  & CF      &  65.43 & 76.54 & \textbf{86.42} & 66.67 && 62.96 \\
  & $\Delta$  & -20.99$^*$ & -9.92$^*$ & -1.23 & \textbf{0.00} && +2.47 \\
\midrule
\multirow{3}{*}{Qwen2.5 32B} 
  & Default & \textbf{92.59} & 91.37 & 88.89 & 70.37 && 75.31 \\
  & CF     & 81.48 & 86.42 & \textbf{90.12} & 74.07 && 74.07 \\
  & $\Delta$   & -11.11$^*$ & -4.94 & \textbf{+1.23} & +4.30 && \textbf{-1.23} \\
\midrule
\multirow{3}{*}{Gemma3 12B} 
  & Default & 90.12 & 87.65 & \textbf{92.59} & 69.14& & 66.67 \\
  & CF      & 72.84 & 75.31 & \textbf{90.12} & 66.67 && 62.96 \\
  & $\Delta$   & -17.32$^*$ & -12.34$^*$ & \textbf{-2.47} & \textbf{-2.47} && -3.71 \\
\midrule
\multirow{3}{*}{Llama3.1 8B} 
  & Default & 86.42 & 87.65 & \textbf{92.59} & 60.49 && 60.49 \\
  & CF      & 48.15 & \textbf{72.84} & \textbf{72.84} & 55.56 && 59.26 \\
  & $\Delta$   & -29.63$^*$ & -14.81$^*$ & -19.75$^*$ & -4.94 && \textbf{-1.23} \\
\bottomrule
\end{tabular}
\caption{
Accuracies on the default and the counterfactual data~(CF) as well as their differences~($\Delta$; 0 is best) on RR. For a robust model, we expect a non-significant difference ($\Delta$) between the Default and CF condition. We mark \textit{brittle} models for which this difference {\it is} significant ($p<0.05$; McNemar's test~\cite{mcnemar1947note}) with an asterisk. The best result per model and metric is marked in bold.
}\label{tab:default_cf_diff}
\end{table*}

\subsection{Neural Approaches}
We compare our neurosymbolic method \textbf{LINC} against three fully neural approaches of varying complexity following~\citet{olausson2023linc}. The input prompt for each model contains 8 in-context learning examples followed by the given premises and conclusion to be evaluated:

\begin{enumerate}
    \item \textbf{Naïve} where we directly prompt an LLM to generate the True/False/Uncertain label.
    The 8-shot examples consist of premise-conclusion pair along with the label.

    \item \textbf{Chain of Thought (CoT)}. Here, our 8-shot examples contain of premises, conclusion and the label, together with a human-created reasoning chain explaining why the label follows from the premises and conclusion pair. We use the reasoning chains given by \citet{olausson2023linc}. We lead the prompt with “Let’s think step by step”.  The output consists of a generated reasoning chain and a final label.

    \item \textbf{ScratchPad}. The LLM is prompted to generate both FOL statements and the True/False/Uncertain label. The scratchpad baseline is included to test whether querying the LLM to generate formal FOL statements can impact its performance in comparision to CoT where we ask for a more ambiguous "reasoning". 
\end{enumerate}

To keep the comparison with LINC fair, for each method we prompt the LLM 10 times to get 10 generated labels and perform majority voting to obtain the final predicted label. We note that LINC is
most likely to benefit from a high number of generations due to its susceptibility to Prover9 errors.

\begin{figure*}
  \centering
  \includegraphics[width=0.8\linewidth]{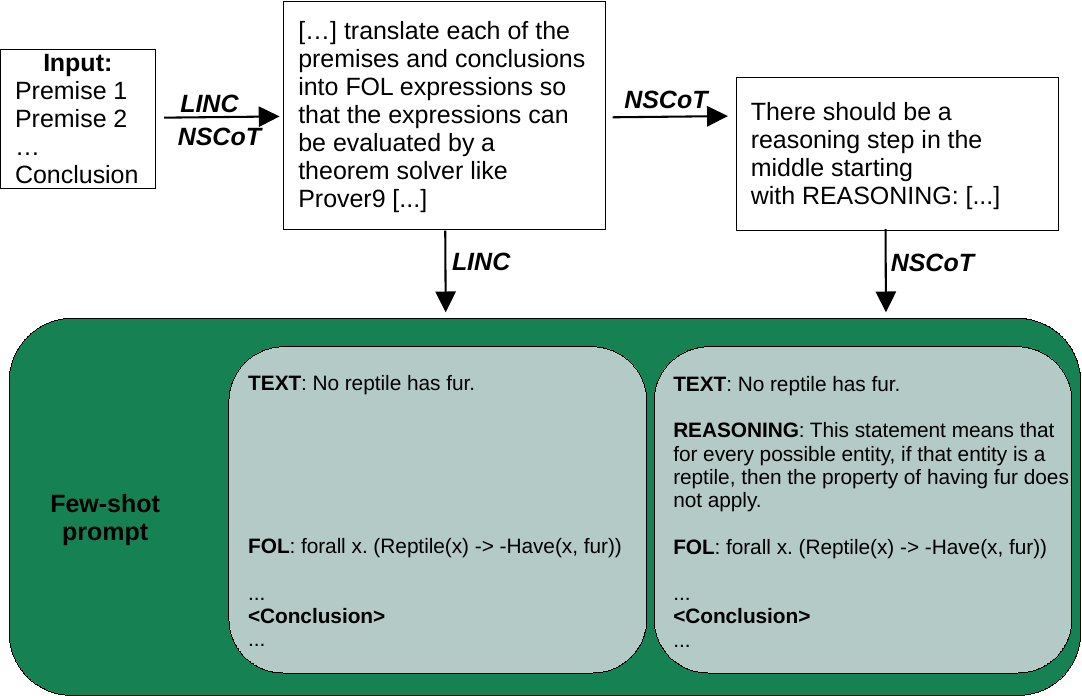} 
  \caption {Comparison of the few-shot prompts in LINC (left) and NSCoT (right). In contrast to LINC, for NSCoT we pass examples that include reasoning chains between the language input and FOL translations; and instruct the model to produce a reasoning chain during generation. After this step, we pass in the generated FOLs to Prover9 for both models.
  }
  \label{fig:fsl-linc-neurocot-nsstep}
\end{figure*}

\subsection{Models}

We test instruction-tuned open-source models from four families and of varying sizes: Mistral0.3 
\citep{jiang2023mistral7b}, Qwen2.5 \citep{Bai_Bai_Chu_Cui_Dang_Deng_Fan_Ge_Han_Huang_etal._2023}, Llama3.1 \citep{dubey2024llama3} and Gemma3 \citep{gemma2024}. These span a representative set of models, with the Qwen and Mistral family are chosen for their focus on reasoning tasks while Llama, and Gemma signify more general-purpose language models.
We test on model sizes between 7 billion and 32 billion parameters where available.

\subsection{Metrics}
To measure robustness to counterfactual perturbations, we simply calculate the difference between the accuracy on the default data and the accuracy on the counterfactual data in the RR dataset.
An ideal model would not be impacted by counterfactual perturbations as these do not impact the logical validity of the inputs. The ideal value of
this accuracy difference is thus zero.

\section{Main Results}

In our main results, we compare the neurosymbolic approach LINC against our three neural baselines on the counterfactually manipulated RR dataset. These results are shown in the left part of \Cref{tab:default_cf_diff}, which shows the accuracies on the default and the counterfactual data along with their differences for these methods. We make a number of observations. 

\paragraph{Robustness}

First, \Cref{tab:default_cf_diff} shows that for LINC, the accuracy differences between the default and the counterfactual data are less than 5\% across all models. This difference is not statistically significant, indicating robustness of LINC against counterfactual manipulation. In contrast, the fully neural methods generally show larger~(over 10\%) and statistically significant accuracy differences. One exception is CoT which shows good robustness albeit inconsistently, with only 3 out of 5 underlying LLMs. These findings suggest that the neurosymbolic LINC approach enhances robustness to counterfactual perturbations, compared to fully neural methods.

\paragraph{Overall performance}

Second, we observe that the neural methods generally outperform LINC in terms of overall performance on both the default and the counterfactual data. 
However, there are exceptions to this trend. For instance, LINC outperforms Naïve on the counterfactual data with Mistral0.3, Qwen2.5 7B, and Llama3.1.
These mixed results suggest that further analysis is necessary. In the next section, we propose a new method designed to improve the overall performance of LINC.

\section{Enhancing LINC with CoT}

We hypothesise that LINC struggles when the natural language statements are convoluted and thus requires more explicit guidance
for NL conversion to FOL statements. Thus, we propose to include an intermediary “reasoning” chain in each few-shot learning example so that the LLM can acquire extra context of how an NL statement should be converted into FOL. We call this approach NSCoT~(short for Neuro-symbolic Chain-of-Thought).

\subsection{Method}
 
We use the ChatGPT o3 reasoning model to generate reasoning chains for our examples, and insert them between the NL and FOL in the prompt. We obtain one reasoning chain for each in-context learning example for a total of 8 reasoning chains~(an abbreviated example, with only 1-shot is shown in \Cref{sec:appendix}). We manually verify the reasoning chains to ensure their correctness. By inserting these reasoning chains, we aim for LLMs not to be confused by examples where the inferred FOL does not directly follow from the text~(cf., \Cref{fig:fsl-linc-neurocot-nsstep}).
To handle Prover9 errors, we follow a similar approach to LINC where we obtain 10 generations and perform majority voting to get the final predicted label.

We note here that the reasoning chain in each in-context learning example is generated using ChatGPT o3 model. This is different from the CoT approach we included in our baseline, where we included human-generated reasoning chains following~\citet{olausson2023linc}. In addition, we prompt NSCoT to perform reasoning in response to each premise individually. Our CoT baseline performs reasoning over all premises at once, in one contiguous block.

\subsection{Evaluation}

We evaluate NSCoT on RR (N=81) under the same conditions as our main model (Table~\ref{tab:default_cf_diff}, right). In addition, we also validate NSCoT, LINC, and selected baseline methods on the full FOLIO validation data set of {\it default} premises (N=204).
This is more than double the size of RR and contains examples that were excluded by \citet{wu2024reasoningreciting} in creating RR.
Due to its increased size and diversity, we expect this dataset to be a more representative testbed for reasoning accuracy on default premises than RR. We report the accuracy numbers in \Cref{tab:methods_accuracy}.

\subsection{Results}

From \Cref{tab:methods_accuracy}, we note several observations. First, the performance of Naïve, CoT, and LINC substantially drops across all models relative to their performance on the default RR data in \Cref{tab:default_cf_diff}. This finding suggests that this larger dataset contains more diverse and challenging examples than RR. Second, Naïve outperforms LINC across three out of five models.\footnote{All models except for Qwen2.5 32B and Gemma3 12B.} Moreover, the CoT baseline consistently outperforms LINC across all five models. This trend is similar to that of \Cref{tab:default_cf_diff} where the baseline methods generally outperform LINC. Third, NSCoT consistently outperforms LINC across all models and outperforms Naïve across 4 out of 5 models.\footnote{Negative case: Gemma3 12B.} This finding highlights the strength of our proposed incorporation of reasoning chains for FOL conversion. That said, NSCoT consistently lags behind the CoT baseline, which underlines the strength of purely neural approaches in terms of overall performance.

\begin{table}[t]
\centering
\begin{tabular}{@{}lrrrr@{}}
\toprule
\textbf{Model} & \textbf{Naïve} & \textbf{CoT} & \textbf{LINC} & \textbf{NSCoT} \\
\midrule
M0.3 7B     & 53.43 & 56.37 & 52.94 & 53.92 \\
Q2.5 7B     & 59.31 & 70.59 & 58.33 & 66.67 \\
Q2.5 32B    & 66.18 & 75.49 & 68.14 & 71.08 \\
L3.1 8B     & 33.33 & 70.59 & 58.33 & 68.14 \\
G3 12B     & 64.22 & 77.45 & 57.35 & 63.24 \\
\bottomrule
\end{tabular}
\caption{Accuracy on the FOLIO validation set~\citep{han-etal-2024-folio} which has more than double the size of RR.
M=Mistral, Q=Qwen, L=Llama, G=Gemma.}
\label{tab:methods_accuracy}
\end{table}

Looking at \Cref{tab:default_cf_diff} to compare LINC and NSCoT on the smaller RR dataset, we observe that, like LINC, NSCoT shows small and non-significant accuracy differences between the default and the counterfactual data for all models. Moreover, these differences are similar in magnitude to those of LINC. This finding suggests that NSCoT is as robust as LINC to counterfactual perturbations.

\begin{figure}[t]
  \includegraphics[width=\columnwidth]{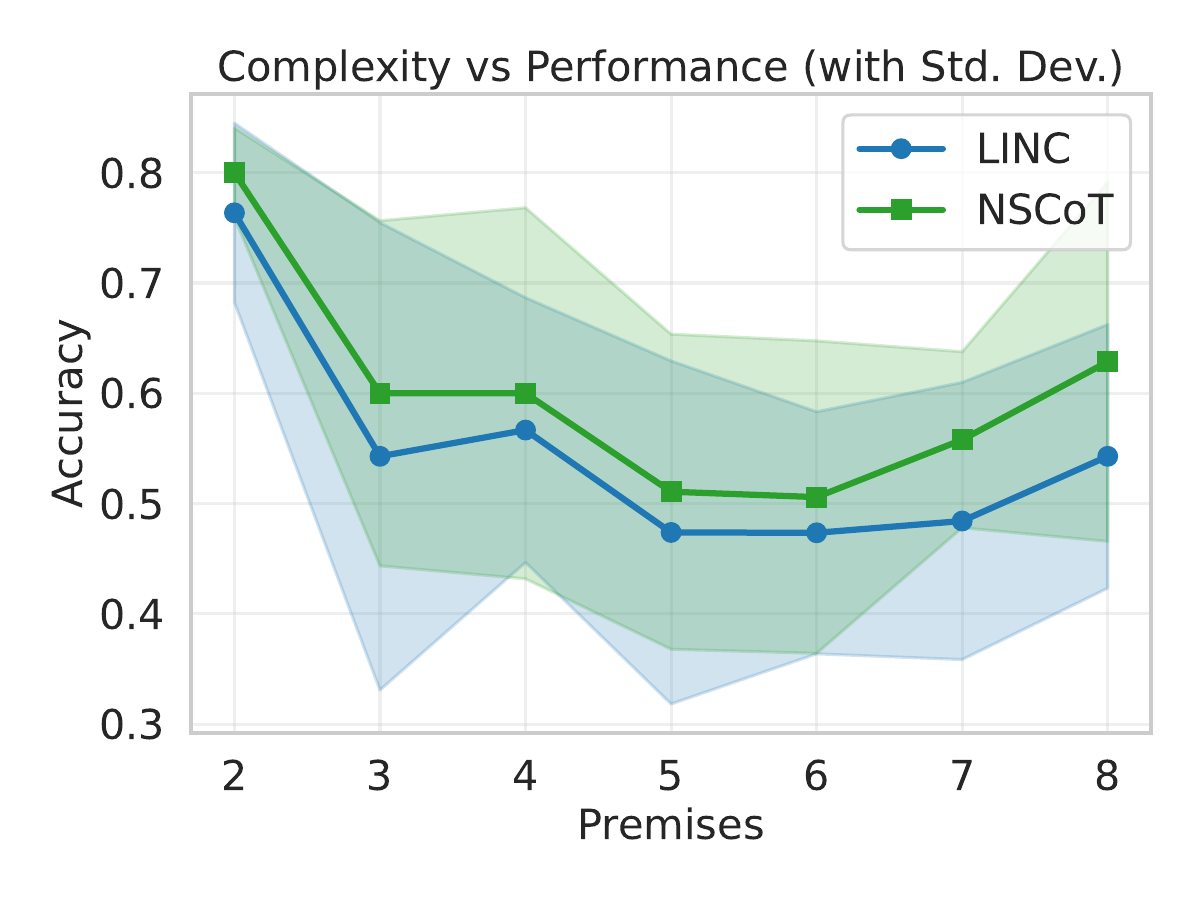}
  \caption{This plot shows the accuracy of LINC (blue) and NSCoT (green) on inputs with different numbers of premises (2 to 8) on the full FOLIO data. The presented results are averaged over all LLMs (as listed in \Cref{tab:methods_accuracy}). LINC suffers a sharper decline in performance than NSCoT. 
   }
  \label{fig:qwen_prem_analysis}
\end{figure}

Instances in the RR data set have an average of 4.3 premises (average length = 283 words), while the FOLIO validation instances have an average of 5.3 premises (average length = 386 words). We checked, using the FOLIO data set, the decline in performance of LINC and NSCoT over instances of increasing complexity, as approximated by the number of premises.
\Cref{fig:qwen_prem_analysis} shows the accuracy of each model on a subset of instances with a fixed number of premises (x-axis; varying from 2 to 8). The result is averaged over all tested LLMs in \Cref{tab:methods_accuracy} (variance is shown as shaded areas). We observe that the gap between NSCoT and LINC increases slightly as the number of premises increases. This suggests that NSCoT effectively leverages the intermediate reasoning step to deal with more complex sets of premises.

From \Cref{tab:default_cf_diff}, we observe that NSCoT outperforms~(or is on par with) LINC on the default data of the smaller RR dataset only with Qwen2.5 32B and Llama 3.1. This is in contrast with the results we observe on the full FOLIO validation set of default premises in \Cref{tab:methods_accuracy} where NSCoT consistently outperforms LINC. It also conflicts with our results in \Cref{fig:qwen_prem_analysis} where NSCoT slightly outperforms LINC for all levels of complexity. We suspect that this discrepancy is due to the sample selection heuristics that \citet{wu2024reasoningreciting} used in creating RR and leave this investigation to future work. We contend that the results in \Cref{tab:methods_accuracy,fig:qwen_prem_analysis} provide stronger evidence for the advantages of NSCoT due to the larger number and more complex examples
in the full FOLIO validation set.

To sum up, our experiments showed that: (1) Neurosymbolic methods outperform purely neural methods in terms of robustness; (2) Pure neural methods, particularly with CoT reasoning, are stronger in terms of accuracy; and (3) the accuracy of neurosymbolic methods can be improved with additional CoT reasoning steps while maintaining strong robustness, albeit not to the level of neural CoT methods. 
The remainder of this paper presents and error analysis and an in-depth discussion of our results, with an eye to future research directions.

\section{Discussion}

\subsection{Class Distributions for Default vs Counterfactual Predictions}
\begin{figure}[t]
  \includegraphics[width=\columnwidth]{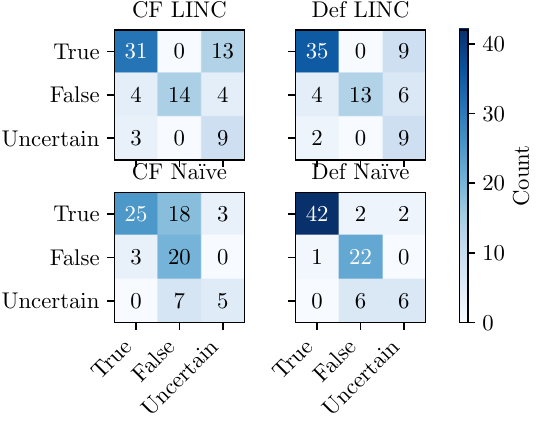}
  \caption{Confusion matrices for the predicted vs gold labels on the CF (left) vs Default (right) versions of RR for LINC (top) and Naïve (bottom). Predicted and ground truth labels are on the x- and y-axis respectively. The underlying LLM is Qwen2.5 7B.}
  \label{fig:cf-vs-def-conf-mats}
\end{figure}

\begin{figure}[t]
  \includegraphics[width=\columnwidth]{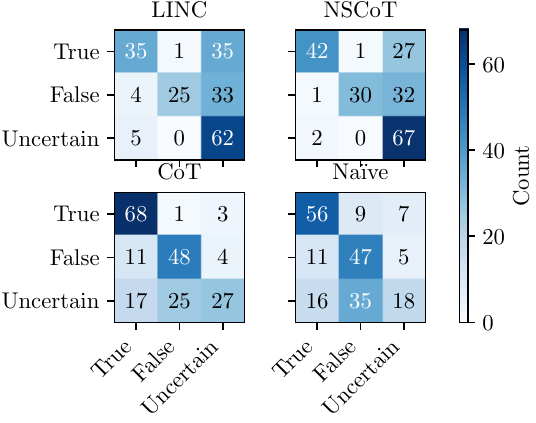}
  \caption{Confusion matrices comparing LINC, NSCoT, CoT and Naïve for the FOLIO validation set. Predicted and ground truth labels are on the x- and y-axis respectively.
  The underlying LLM is Qwen2.5 7B.}
  \label{fig:full-data-conf-mats}
\end{figure}

Recall that the ground truth label distribution in the default and counterfactual (CF) versions of RR are identical, as the perturbations had no bearing on the logical conclusion of the premises. We thus compare the confusion matrices between predicted and ground truth class distributions of several models. We start by inspecting the predicted class distribution shift for LINC and the Naïve method on RR (\Cref{fig:cf-vs-def-conf-mats}), and subsequently compare the label distributions of all tested methods on the larger FOLIO validation data set (\Cref{fig:full-data-conf-mats}). All results are based on Qwen2.5 7B.

\Cref{fig:cf-vs-def-conf-mats} shows that while LINC maintains a nearly identical confusion matrix profile across both settings, the Naïve method shows a noticeable shift in the distribution of labels. Particularly, around 20\% of samples flip from the True class to the False class, leading to a substantial reduction in accuracy. This behavior reflects the Naïve model’s tendency to rely on surface-level token associations, which collapse when predicates or constants are perturbed. In contrast, LINC’s symbolic pipeline ensures that perturbations are more likely to lead to either consistent or \textit{Uncertain} predictions.

\Cref{fig:full-data-conf-mats} compares the confusion matrices for Naive, CoT, NSCoT and LINC based on the Qwen7B instruction fine-tuned model and on the larger FOLIO validation data. 
We can observe that both the CoT and Naïve models show a higher False Negative (FN) rate for \textit{Uncertain} class instances i.e. {Naïve and CoT} methods both tend to under-predict \textit{Uncertain}. 
This suggests that neural methods {overfit to surface regularities}, confidently outputting categorical answers even when evidence is ambiguous. 

In contrast, the neurosymbolic methods ({LINC and NSCoT}) produce more {\it Uncertain} predictions. However, this comes with a trade-off: some of these are false positives because the LLM produces predicates with overlapping meaning, and Prover9 as a symbolic solver cannot detect this, thus predicting Uncertain for otherwise resolvable cases. Overlapping meaning refers to instances containing distinct predicates with shared denotation that form a disconnect in the logical flow. For example, if the LLM output contains the predicates ``Dog'' and ``CuteDog'' then Prover9 will not be able to be resolve them~(i.e. that ``CuteDog'' implies ``Dog''), causing the logical reasoning process to fail. A more elaborate example is included in \Cref{appendix_a2}. 
The quantity of these errors is captured by the Uncertain false positives in the respective confusion matrices of LINC and NSCoT (Figure~\ref{fig:full-data-conf-mats}). This error class is an instance of the more general issue of implicit information in NL statements (e.g., that ``CuteDog'' implies ``Dog''), which has also been noted in prior work~\citet{olausson2023linc}.

\subsection{Error analysis for NSCoT and LINC}
\label{sec:error_analysis}

To better understand the FOL conversion errors of both LINC and NSCoT, we manually classified the observed errors on all the examples from the FOLIO validation set where the methods generated FOL statements that did not comply with the Prover9 syntax. Here, "error" is generations in which Prover9 was not able to resolve the given FOL statements due to the specified error classes. These were 341 cases for LINC and 366 cases for NSCoT, out of a total of 2040 queries (204 examples $\times$ 10 generations).
We found two common classes of erros. The first are \textit{arity mismatches} where predicates are used with inconsistent numbers of arguments across premises in the same instance (e.g. Likes(x,y) vs. Likes(x)). 
The second common error class pertains to \textit{unexpected tokens}.
This typically arises based on malformed or incomplete FOL strings (e.g., missing parentheses, unbalanced connectives), which cause Prover9 to throw parsing errors.

\begin{figure}[t]
  \includegraphics[width=\columnwidth]{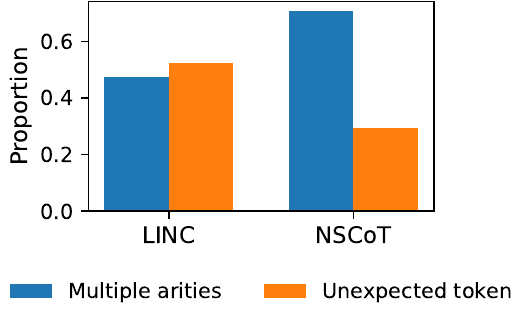}
  \caption{
  Proportion of the two most common FOL conversion errors of both LINC and NSCoT: arity mismatch and unexpected token. The underlying LLM is Qwen2.5 7B.
  }
  \label{fig:error-analysis}
\end{figure}

\Cref{fig:error-analysis} shows the relative prevalence of both error classes for LINC and NSCoT with Qwen2.5 7B. NSCoT produces more arity mismatch errors compared to LINC. However, LINC produces more unexpected token errors than NSCoT.

We conducted preliminary tests with a verification module to refine generations which did not execute due to Prover9 errors on both methods. In this setup, the Prover9 error messages and a few examples of common syntax corrections were put into a new prompt, and the model was re-queried in a loop until the FOL expressions executed successfully or a maximum of 3 retries was reached. However we had similar findings to \citet{pan2023logiclm} in which they showed that the execution rate of the symbolic prover increases using a refiner but at the same time the accuracy decreases due to more semantic errors.

\subsection{Faithfulness of CoT}

While CoT outperforms NS methods in terms of accuracy and achieves comparable robustness on three out of six models (\Cref{fig:cf-vs-def-conf-mats}), the underlying reasoning can be “unfaithful”, introducing hallucinated steps or logical inconsistencies. CoT's high false negative rate for the {Uncertain} label fits with a concern about unfaithfulness: the models CoT text output is not actually representing internal logical reasoning, but rather reproducing the most frequently observed labels during training which was either a True or False label.

The limits of CoT prompting have come under scrutiny by \citet{lanham2023faithfulness} who showed that CoT rationales often reflect post-hoc justifications rather than the true decision process of the model. They modified the generated CoT reasoning such as making it incomplete, masking some tokens, or introducing some mistakes. They then re-queried the model with the modified reasoning and found that the model still gives the correct output, suggesting that the output does not actually depend on the printed reasoning chains, and that the reasoning chains are produced post-hoc after the final class prediction has already been computed.

An important and intriguing path for future work could be to leverage this method to test whether CoT faithfulness changes between CF and default samples. If the method genuinely reasons over the given input, we expect to not observe a drop in faithfulness since the complexity of the problems stays fixed due to the design of the CF examples.

\section{Conclusion}

We have presented the first rigorous comparison of strong neural methods with the neurosymbolic method LINC --- which combines LLM-based natural language to FOL parsing with an FOL solver --- on the task of logical reasoning. We showed that while LINC shows stronger robustness results, it falls short of the neural methods in terms of performance. We then extended LINC with CoT reasoning steps showing that reasoning accuracy is enhanced while maintaining robustness. However, the fully neural methods still achieve the strongest results based on performance.

This paper addresses the timely and relevant question of neurosymbolic approaches in AI which are desirable due to a promise of a decreased carbon footprint due to the outsourcing of part of the reasoning to efficient external modules (such as logical reasoners). Furthermore, neurosymbolic approaches promise a tighter control on interpretability and faithfulness of the results. Our results present a step in this direction, by carefully evaluating NS methods on logical reasoning and proposing steps for future research.

We note that further optimization is most critical for improving the \textbf{accuracy} of the neurosymbolic methods, rather than robustness, as the main source of performance degradation comes from inconsistent NL–FOL translations. Optimizing this stage would therefore allow the symbolic reasoning component compute on correct logical forms, yielding more faithful and accurate deductions.

\section*{Limitations}
Our work is limited to one type of reasoning namely first-order logic and should in the future be expanded to tasks such as math word problems~\citep{huang2025mathperturb} (using SymPy as a symbolic solver), coding, and planning. There is a need for systems which are more robust to variation and faithful under hard problems. Other limitations of this study include the small test data size and exclusion of larger-sized models due to computational resources.
\section*{Acknowledgments}
This research was supported by The University of Melbourne’s Research Computing Services and the Petascale Campus Initiative.

\bibliography{custom}

\newpage
\appendix

\section{Appendix}
\label{sec:appendix}

\subsection{1-shot example for NSCoT}

\noindent
\fbox{%
\parbox{0.5\textwidth}{%

\small
\textbf{Task.} The following is a first-order logic (FOL) problem. The problem is to determine whether the conclusion follows from the premises. The premises are given in the form of a set of first-order logic sentences. The conclusion is given in the form of a single first-order logic sentence. The task is to translate each of the premises and conclusions into FOL expressions so that the expressions can be evaluated by a theorem solver like Prover9 to determine whether the conclusion follows from the premises. There should be a reasoning step in the middle starting with REASONING:..  The output should follow format TEXT:  REASONING:  FOL:  TEXT:  REASONING:  FOL:...Expressions should adhere to the format of the Python NLTK package logic module.Remember to never the FOL predicates too long or overlapping meanings with other predicates.  Also you must not have empty predicates like Conductor(). Symbols like <, >, = are not allowed.

\vspace{0.4em}

\textbf{Premises.}
\begin{itemize}
  \item All dispensable things are environment-friendly.
  \item All woodware is dispensable.
  \item All paper is woodware.
  \item No good things are bad.
  \item All environment-friendly things are good.
  \item A worksheet is either paper or is environment-friendly.
\end{itemize}

\textbf{Conclusion.} A worksheet is not dispensable.

\vspace{0.4em}
\textbf{Evaluate.}

\textbf{TEXT:} All dispensable things are environment-friendly.\\
\textbf{REASONING:} We universally quantify over every entity, asserting that being dispensable implies being environment-friendly.\\
\textbf{FOL:} \text{all x. (Dispensable(x) -> EnvironmentFriendly(x))}

\medskip
\textbf{TEXT:} All woodware is dispensable.\\
\textbf{REASONING:} Every instance of woodware falls under the dispensable category.\\
\textbf{FOL:} \text{all x. (Woodware(x) -> Dispensable(x))}

\medskip
\textbf{TEXT:} All paper is woodware.\\
\textbf{REASONING:} Paper is a subset of woodware, so paper implies woodware.\\
\textbf{FOL:} \text{all x. (Paper(x) -> Woodware(x))}

\medskip
\textbf{TEXT:} No good things are bad.\\
\textbf{REASONING:} For every x, if x is good then x cannot be bad.\\
\textbf{FOL:} \text{all x. (Good(x) -> -Bad(x))}

\medskip
\textbf{TEXT:} All environment-friendly things are good.\\
\textbf{REASONING:} Environment-friendly implies good.\\
\textbf{FOL:} \text{all x. (EnvironmentFriendly(x) -> Good(x))}

\medskip
\textbf{TEXT:} A worksheet is either paper or is environment-friendly.\\
\textbf{REASONING:} Worksheet must be in exactly one of those two categories.\\
\parbox{0.95\linewidth}{\raggedright\textbf{FOL:} \texttt{((Paper(Worksheet) \& -EnvironmentFriendly(Worksheet)) | (-Paper(Worksheet) \& EnvironmentFriendly(Worksheet)))}}

\medskip
\textbf{TEXT:} A worksheet is not dispensable.\\
\textbf{REASONING:} We assert the negation of dispensability for the specific constant \text{Worksheet}.\\
\textbf{FOL:} \text{-Dispensable(Worksheet)}
}%
}

\subsection{Comparison between consistent and ambiguous FOL generations}\label{appendix_a2}

\begin{table*}
\centering
\renewcommand{\arraystretch}{0.95}
\setlength{\tabcolsep}{2.5pt}

\begin{tabular}{p{0.025\linewidth} | p{0.55\linewidth} | p{0.4\linewidth}}
\textbf{\#} &
\textbf{Generation 1 (Ambiguous)} &
\textbf{Generation 2 (Consistent)} \\ \hline

1 &
$\forall x,y.\; (\textcolor{teal}{\mathrm{WildTurkey}}(x) \wedge \textcolor{teal}{\mathrm{WildTurkey}}(y) \wedge
(x = \mathrm{EasternWildTurkey} \vee x = \mathrm{OsceolaWildTurkey} \vee 
x = \mathrm{GouldsWildTurkey} \vee x = \mathrm{MerriamsWildTurkey} \vee 
x = \mathrm{RioGrandeWildTurkey} \vee x = \mathrm{OcellatedWildTurkey}) \wedge 
x \neq y \rightarrow \bot)$
&
$\forall x.\; (\textcolor{teal}{\mathrm{WildTurkey}}(x) \rightarrow
(\mathrm{EasternWildTurkey}(x) \vee
\mathrm{OsceolaWildTurkey}(x) \vee
\mathrm{GouldsWildTurkey}(x) \vee
\mathrm{MerriamsWildTurkey}(x) \vee
\mathrm{RioGrandeWildTurkey}(x) \vee
\mathrm{OcellatedWildTurkey}(x)))$ \\ \hline

2 & $\neg\textcolor{red}{\mathrm{WildTurkeyType}}(\mathrm{Tom}, \mathrm{EasternWildTurkey})$ & $\neg\mathrm{EasternWildTurkey}(\mathrm{Tom})$ \\ \hline
3 & $\neg\textcolor{red}{\mathrm{WildTurkeyType}}(\mathrm{Tom}, \mathrm{OsceolaWildTurkey})$ & $\neg\mathrm{OsceolaWildTurkey}(\mathrm{Tom})$ \\ \hline
4 & $\neg\textcolor{red}{\mathrm{WildTurkeyType}}(\mathrm{Tom}, \mathrm{GouldsWildTurkey}) \wedge 
\neg\textcolor{red}{\mathrm{WildTurkeyType}}(\mathrm{Tom}, \mathrm{MerriamsWildTurkey}) \wedge 
\neg\textcolor{red}{\mathrm{WildTurkeyType}}(\mathrm{Tom}, \mathrm{RioGrandeWildTurkey})$ &
$\neg\mathrm{GouldsWildTurkey}(\mathrm{Tom}) \wedge
\neg\mathrm{MerriamsWildTurkey}(\mathrm{Tom}) \wedge
\neg\mathrm{RioGrandeWildTurkey}(\mathrm{Tom})$ \\ \hline
5 & $\textcolor{teal}{\mathrm{WildTurkey}}(\mathrm{Tom})$ & $\textcolor{teal}{\mathrm{WildTurkey}}(\mathrm{Tom})$ \\ \hline
6 & $\textcolor{red}{\mathrm{WildTurkeyType}}(\mathrm{Tom}, \mathrm{OcellatedWildTurkey})$ & $\mathrm{OcellatedWildTurkey}(\mathrm{Tom})$ \\ \hline

\multicolumn{1}{c|}{} &
\textit{\footnotesize Problem: The predicate \textcolor{red}{WildTurkeyType} is never linked to 
\textcolor{teal}{WildTurkey}, creating ambiguity between types and individuals.} &
\textit{\footnotesize Correct: All predicates share the same unary form, so Tom’s type is inferred successfully.}
\\
\hline
\end{tabular}

\caption{The correct inference is \textbf{True}, but ambiguous predicate names in ``Generation 1'' lead to \textbf{Uncertain}. 
We compare ambiguous (left) and consistent (right) FOL statements with predicate numbering.
\textcolor{red}{Red} indicates inconsistent predicate forms causing uncertainty; 
\textcolor{teal}{Teal} indicates consistent unary naming that yields a correct inference.}
\label{tab:example_uncertain_fp}
\end{table*}

\end{document}